\documentclass[twoside,11pt]{article}

\usepackage{jmlr2e}
\usepackage[figuresright]{rotating}
\usepackage{float}
\usepackage{amsmath}
\usepackage{multirow}
\usepackage{threeparttable}

\jmlrheading{}{}{}{}{}{}{Amir M. Mir, Mahdi Rahbar, and Jalal A. Nasiri}

\ShortHeadings{LIBTwinSVM: A Library for Twin Support Vector Machines}{Mir, Rahbar, Nasiri}
\firstpageno{1}

\newcommand{\bsym}[1]{\mathbf{#1}}

\begin{document}

\title{LIBTwinSVM: A Library for Twin Support Vector Machines}

\author{\name Amir M. Mir\textsuperscript{\dag\ddag} \email a.mir@iau-tnb.ac.ir \\
	   \name Mahdi Rahbar\textsuperscript{\dag\ddag} \email mehdi.rahbar@iau-tnb.ac.ir \\
	   \name Jalal A. Nasiri\textsuperscript{\ddag} \email j.nasiri@irandoc.ac.ir \\
       \addr \textsuperscript{\dag}Faculty of Electrical and Computer Engineering, Islamic Azad University, North Tehran Branch, Tehran, Iran\\
       \addr \textsuperscript{\ddag}Iranian Research Institute for Information Science and Technology (IranDoc), Tehran, Iran}

\editor{}

\maketitle

\begin{abstract}%   <- trailing '%' for backward compatibility of .sty file
This paper presents LIBTwinSVM, a free, efficient, and open source library for Twin Support Vector Machines (TSVMs). Our library provides a set of useful functionalities such as fast TSVMs estimators, model selection, visualization, a graphical user interface (GUI) application, and a Python application programming interface (API). The benchmarks results indicate the effectiveness of the LIBTwinSVM library for large-scale classification problems. The source code of LIBTwinSVM library, installation guide, documentation, and usage examples are available at \href{https://github.com/mir-am/LIBTwinSVM}{https://github.com/mir-am/LIBTwinSVM}.
\end{abstract}


\section{Introduction}\label{sec:1}
%Support Vector Machine (SVM) is a powerful classification algorithm \citep{vapnik1995}. Its main idea is to find the optimal separating hyperplane with the largest margin. Due to the SVM's great generalization ability, It has been successfully applied to a wide variety of applications including arrhythmia classification \citep{nasiri2009}, text categorization \citep{lee2012}, and spam detection \citep{olatunji2019}. However, SVM has high computational time for large-scale datasets. Because it solves a large Quadratic Programming Problem (QPP) which has a time complexity of $\mathcal{O}(n^3)$. On the basis of SVM, scholars have proposed novel classifiers in the past decade \citep{chauhan2018}.

Twin Support Vector Machine (TSVM) is an extension of the Support Vector Machine (SVM), which was proposed by \citet{Khemchandani2007}. TSVM does binary classification using two non-parallel hyperplanes. Each of which is as close as possible to the samples of its own class and far from the samples of the other class. The two non-parallel hyperplanes are obtained by solving two smaller-sized Quadratic Programming Problems (QPPs). This makes the learning speed of TSVM four times faster than that of SVM in theory \citep{Khemchandani2007}. Over the past decade, new extensions of TSVM were introduced. For instance, \citet{kumar2009} proposed Least Squares Twin Support Vector Machine (LSTSVM). Its central idea is similar to TSVM. However, LSTSVM solves two systems of linear equations as opposed to solving two QPPs.

Aside from new research works on the theory of SVM, numerous free software packages and libraries have released for SVM, which include but not limited to, LIBLINEAR \citep{fan2008}, LIBSVM \citep{chang2011}, and ThunderSVM \citep{wen2018}. However, there are not many free and open source software packages available for TSVM. Recently, \citet{ltsvm2019} released LightTwinSVM program which is a simple and fast implementation of standard TSVM. It is a light-weight and small program that only provides a command-line tool for classification and model selection.

Motivated by the above discussion, we present LIBTwinSVM which is a free and open source library for Twin Support Vector Machines. It is licensed under the terms of GNU General Public License v3.0, which means this library can be used for both academic and commercial purposes. Different from other SVM and TSVM software packages, our library has a simple Graphical User Interface (GUI) to help users employ the functionalities of the library without writing code. In addition, a Python application programming interface (API) was provided. The LIBTwinSVM library can be used on Linux, Windows, and Macintosh operating systems. This library can be found on GitHub at \href{https://github.com/mir-am/LIBTwinSVM}{https://github.com/mir-am/LIBTwinSVM}.

%The key features of LIBTwinSVM are summarized as follows:
%\begin{itemize}
%	\item A simple Graphical User Interface (GUI) to help users employ the functionalities of LIBTwinSVM without writing code.
%	\item Provides fast implementation of standard TSVM and LSTSVM classifiers.
%	\item Has dataset processing module with normalization and shuffling capabilities.
%	\item To solve optimization problems of TSVM, clipping dual coordinate descent (clipDCD) algorithm \citep{peng2014} was improved and implemented in C++.
%	\item Supports linear, Gaussian (RBF), and Rectangular kernel.
%	\item To solve multi-class classification problems, both One-vs-All and One-vs-One schemes are supported.
%	\item Provides model selection and evaluation facilities such as K-fold cross-validation, train/test split, and grid search.
%	\item Detailed classification results can be saved in a spreadsheet file for further analysis.
%	\item A visualization tool to show high-quality decision boundaries for both binary and multi-class TSVM classifiers with linear and non-linear kernels.
%	\item  The best-fitted model can be saved on the disk. In addition, pre-trained models can be loaded and evaluated on test samples.
%\end{itemize}

%This paper is organized as follows.

\section{Twin Support Vector Machine}\label{sec:2}
In this section, we briefly review TSVM and its optimization problems. The linear TSVM finds a pair of non-parallel hyperplanes which are represented by ${\bsym{x}^T}{\bsym{w}_{1}}+{{b}_{1}}=0$ and ${\bsym{x}^T}{\bsym{w}_{2}}+{{b}_{2}}=0$, where $\bsym{x}, \bsym{w}_{1}, \bsym{w}_{2} \in \mathbb{R}^{d}$ and $b_{1},b_{2} \in \mathbb{R}$. Let the positive and negative samples be denoted by matrix $\bsym{A} \in \mathbb{R}^{n_1 \times d}$ and $\bsym{B} \in \mathbb{R}^{n_2 \times d}$, respectively. To obtain the two non-parallel hyperplanes, TSVM solves two primal optimization problems with objective function corresponding to one class and constrains corresponding to other class.
\begin{align}
\label{eq:2}
\begin{split}
\mathop{{ min}}\limits_{\bsym{w}_{1} ,b_{1}} \qquad & \frac{1}{2}{{\left\| \bsym{A}\bsym{w}_{1}+\bsym{e}_{1}{{b}_{1}} \right\|}^{2}}+{c}_{1}\bsym{e}_{2}^{T}\bsym{\xi} \\
\textrm{s.t. } \qquad & -(\bsym{B}\bsym{w}_{1}+\bsym{e}_{2}{b}_{1})+\bsym{\xi} \ge \bsym{e}_{2}\text{ },\bsym{\xi} \ge 0
\end{split} \\
\label{eq:3}
\begin{split}
\mathop{{ min}}\limits_{\bsym{w}_{2} ,b_{2}} \qquad & \frac{1}{2}{{\left\| \bsym{B}\bsym{w}_{2}+\bsym{e}_{2}{b}_{2} \right\|}^{2}}+{c}_{2}\bsym{e}_{1}^{T}\bsym{\eta} \\
\textrm{s.t. } \qquad & (\bsym{A}\bsym{w}_{2}+\bsym{e}_{1}{{b}_{2}})+\bsym{\eta} \ge \bsym{e}_{1}\text{ },\eta\ge 0
\end{split}
\end{align}
\noindent where $c_1$ and $c_2$ are positive penalty parameters, $\xi_{1}$ and $\xi_{2}$ are slack vectors, $e_1$ is the column vectors of ones of $n_{1}$ dimensions and $e_{2}$ is the column vectors of ones of $n_{2}$ dimensions. Similar to SVM, the QPPs (\ref{eq:2}) and (\ref{eq:3}) are converted to dual problems, which are easier to solve and has the advantage of bounded constraint. A new sample $\bsym{x} \in \mathbb{R}^{d}$ is assigned to the class $+1$ or $-1$ depending on which of the two hyperplanes lies closest to the sample in terms of perpendicular distance. TSVM was also extended to handle non-linear problems by using two non-parallel kernel generated-surfaces \citep{Khemchandani2007}.

\section{Library Description}\label{sec:3}
\begin{figure}[t]
	\centering
	\includegraphics[width=1.0\columnwidth]{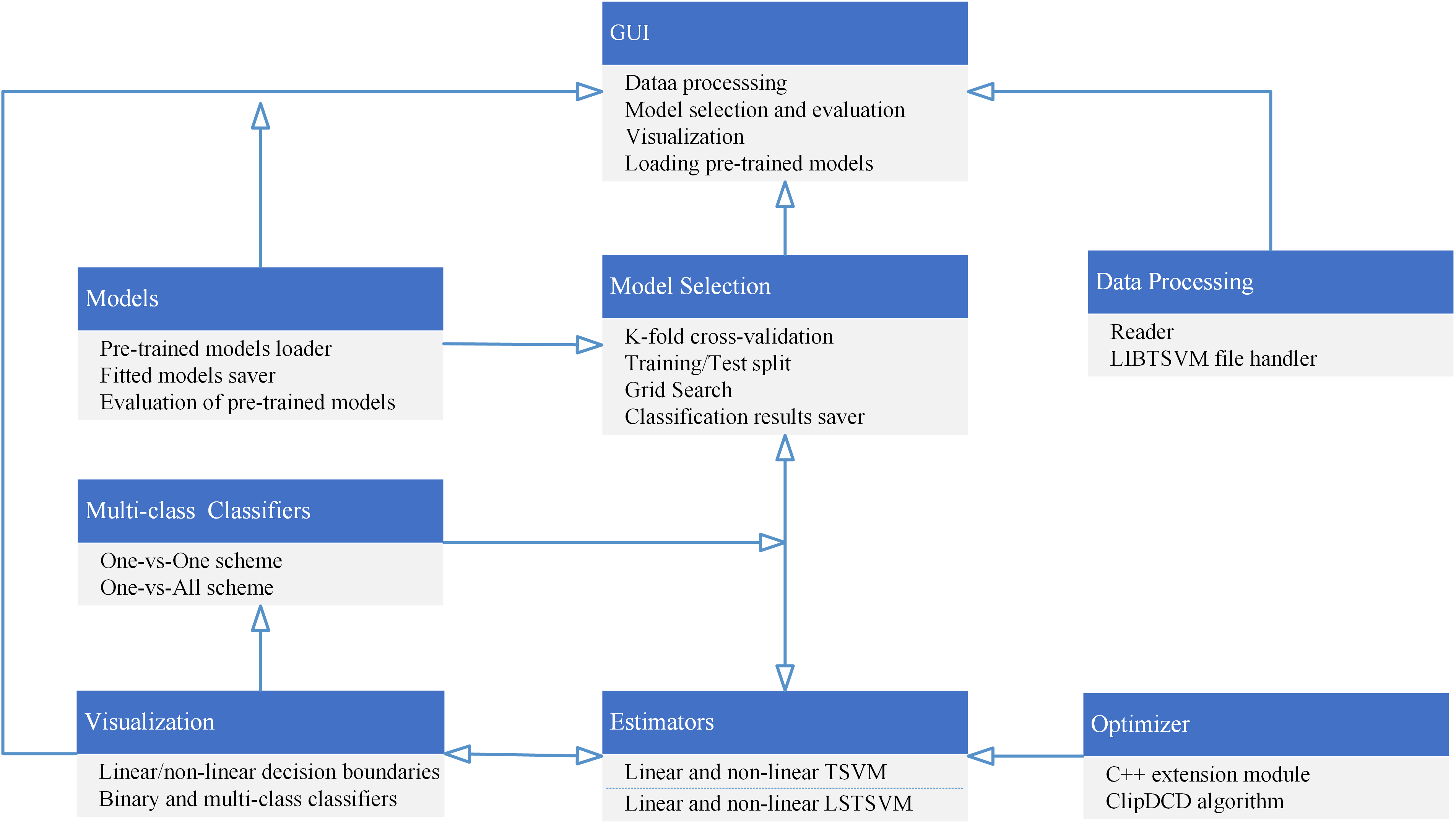}
	\caption{The main components of the LIBTwinSVM library and their relation.}
	\label{fig:1}
\end{figure}

In this section, we discuss the design and implementation of the LIBTwinSVM library. The library was developed in a modular manner, which makes debugging, testing and maintenance easier. Figure \ref{fig:1} shows the main components of the LIBTwinSVM library and their relation. Most SVM and TSVM packages have a command-line interface including LIBSVM, ThunderSVM, and LightTwinSVM. However, our library provides a user-friendly graphical user interface (GUI) which makes usage of library functionalities easier. LIBTwinSVM also consists of fast implementation of standard TSVM and LSTSVM classifier. On the other hand, LightTwinSVM program only provides an implementation of standard TSVM. To solve multi-class classification problems, popular multi-class schemes including One-vs-All (OVA) and One-vs-One (OVO) are provided.

\begin{figure}[t]
	\centering
	\includegraphics[width=1.0\columnwidth]{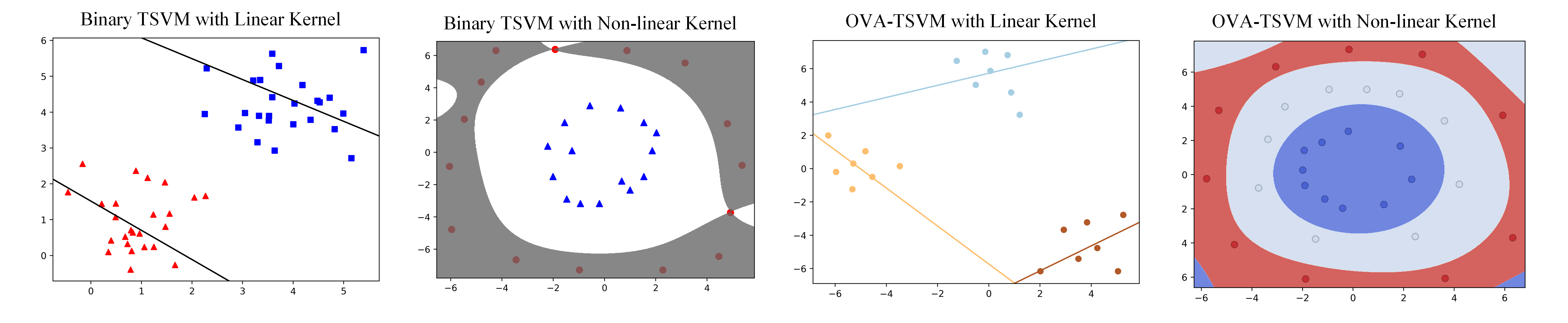}
	\caption{A sample output of visualization capabilities of LIBTwinSVM.}
	\label{fig:3}
\end{figure}

LIBTwinSVM library contains a visualization component, which can draw linear and non-linear decision boundaries for both binary and multi-class TSVM-based estimators. Figure \ref{fig:3} indicates a sample output of the visualization capabilities of LIBTwinSVM. Moreover, the models component of our library allows users to save the best-fitted estimator on the disk. A pre-trained estimator can also be loaded and evaluated on test samples.

%\begin{figure}[t]
%	\centering
%	\includegraphics[width=1.0\columnwidth]{figs/GUI}
%	\caption{The GUI of the LIBTwinSVM library.}
%	\label{fig:2}
%\end{figure}

\subsection{Implementation}\label{sec:2:1}
In this subsection, we give the implementation details of the LIBTwinSVM library. Our library is mostly written in Python 3. It relies completely on open source and free Python packages which are explained as follows. We used Numerical Python (NumPy) for linear algebra operations \citep{walt2011}. For model selection and evaluation, scikit-learn was employed \citep{pedregosa2011}. Pandas package was used for reading and processing datasets. Moreover, the visualization capabilities of our library were implemented using Matplotlib package \citep{hunter2007}. We developed the GUI application of the library using PyQT, which is a powerful and cross-platform GUI package. For saving and loading pre-trained models, we employed Joblib\footnote{https://joblib.readthedocs.io/en/latest/} package.

The optimizer component is the performance-critical part of the LIBTwinSVM library. Therefore, we implemented clipDCD optimizer \citep{peng2014} in C++ for solving dual optimization problems efficiently. As suggested by \citet{peng2014}, we used a caching technique to store the major computations with space cost $\mathcal{O}(n)$. Moreover, our implementation of clipDCD optimizer uses the Armadillo \citep{Sanderson2016}, which is a fast linear algebra library for C++. Since the optimizer component is entirely written in C++, we used Cython \citep{behnel2011} to generate a C++ extension module for Python. Different from the LightTwinSVM program, the implementation of the optimizer component avoids memory copies. This improves substantially the speed and memory consumption of the TSVM estimator. Additionally, the peak memory consumption of the TSVM estimator is further reduced by changing the order of computation in the training stage. The decision function of TSVM-based estimators is evaluated in parallel by using NumPy arrays rather than a for-loop.

\subsection{API}\label{sec:2:2}
This subsection describes the design of the Python API of the LIBTwinSVM library. The main design goal of the API is simplicity and ease of use. Therefore, API functionalities can be used with the basic knowledge of the Python language. \texttt{TSVM} and \texttt{LSTSVM} are binary TSVM-based estimators in the API of our library. Inspired by the design of the scikit-learn package, the interface of all the estimators provides \texttt{fit()} and \texttt{predict()} methods. Similar to the scikit-learn's interface, the library's estimators accept both a Python list and NumPy arrays. For ease of use, the estimators can be initialized with default hyper-parameters. \texttt{OneVsOneClassifier} and \texttt{OneVsAllClassifier} are multi-class classifiers which implemented as \textit{meta-estimators}. They take as input a binary TSVM-based estimator. Moreover, the optimizer component provides a \texttt{clipdcd.optimize()} method for solving dual QPPs. The API reference and its usage examples can be found at \href{https://libtwinsvm.readthedocs.io}{https://libtwinsvm.readthedocs.io}.

\section{Benchmarks}\label{sec:4}
In this section,  we conduct an experiment to show the computational efficiency of LIBTwinSVM library for large-scale datasets. The experiment was carried out on a PC with Intel Core i7 6700K CPU, 32.0 GB of RAM, and Ubuntu 16.04.6 LTS. For this experiment, we used David Musicant's NDC Data Generator \citep{musicant1998ndc} to generate binary classification problems with 32 features. It should be noted that the multi-core version of LIBLINEAR (v2.30)\footnote{https://www.csie.ntu.edu.tw/~cjlin/libsvmtools/multicore-liblinear/} is tested on NDC datasets. The hyper-parameters of the estimators were set to default values (i.e. $c_{1}=1, c_{2}=1$). The linear kernel was also used.  

Table \ref{tab:1} shows the comparison of LIBTwinSVM with other implementations in terms of learning and prediction speed. The learning and prediction speed of the LIBTwinSVM's TSVM estimator is much faster than that of LightTwinSVM. Because we improved the implementation of the TSVM estimator (as discussed in Subsection \ref{sec:2:1}). For instance, the training speed of the library's TSVM estimator is about 6 times faster than the TSVM estimator of LightTwinSVM on NDC-50K dataset. Thanks to the Least Squares extension of TSVM classifier, our library can handle large-scale datasets with 5 million samples whereas LightTwinSVM is not suitable for datasets with more than 50,000 samples. Moreover, the learning speed of LIBTwinSVM with LSTSVM classifier is significantly faster than that of LIBLINEAR, which is a state-of-the-art implementation of linear SVM.

%\begin{table}[!t]
%	\small
%	\centering
%	\begin{threeparttable}
%		\resizebox{\textwidth}{!}{\begin{tabular}{|c|c|c|c|c|c|c|}
%			\hline
%			\multirow{3}{*}{Datasets (no. samples) }& \multicolumn{2}{c|}{LightTwinSVM} & \multicolumn{4}{c|}{LIBTwinSVM} \\ \cline{2-3} \cline{4-7}
%			&\multirow{2}{*}{Train time (s)}  & \multirow{2}{*}{Test time (s)} & \multicolumn{2}{c|}{TSVM} & \multicolumn{2}{c|}{LSTSVM} \\ \cline{4-5} \cline{6-7}
%			& & & Train time (s) & Test time (s) & Train time (s) & Test time (s) \\
%			\hline
%		    NDC-5K (5,000) & 0.57 & 0.012 &  0.63 & 0.00037 & 0.0025 & 0.00032 \\
%			NDC-10K (10,000) & 2.21 & 0.024 &  2.47 & 0.00062 & 0.0047 & 0.00067 \\
%			NDC-25K (25,000) & 13.99 & 0.055 & 15.69 & 0.0016 & 0.013 & 0.0019 \\
%			NDC-50K (50,000) & 56.39 & 0.087 & 63.85 & 0.062 & 0.021 & 0.0022 \\
%			NDC-1l (100,000) & \tnote{a} & \tnote{a} & \tnote{a} & \tnote{a} & 0.038 & 0.0047 \\
%	        NDC-5l (500,000) & \tnote{a} & \tnote{a} & \tnote{a} & \tnote{a} & 0.20 & 0.041 \\
%	        NDC-1m (1,000,000) & \tnote{a} & \tnote{a} & \tnote{a} & \tnote{a} & 0.38 & 0.074 \\
%			\hline
%		\end{tabular}}
%		\begin{tablenotes}
%			\item[a] The experiment ran out of memory.
%		\end{tablenotes}
%	\end{threeparttable}
%\caption{A performance comparison between LightTwinSVM and LIBTwinSVM in terms of learning and prediction speed.}
%\label{tab:1}
%\end{table}

%\begin{table}[!t]
\begin{sidewaystable*}
	\centering
	\begin{threeparttable}
		\resizebox{\textwidth}{!}{\begin{tabular}{|c|c|c|c|c|c|c|c|c|c|c|c|c|}
				\hline
				\multirow{3}{*}{Datasets (no. samples) }& \multicolumn{3}{c|}{LIBLINEAR} &\multicolumn{3}{c|}{LightTwinSVM} & \multicolumn{6}{c|}{LIBTwinSVM} \\ \cline{2-4} \cline{5-7} \cline{7-13}
				& \multirow{2}{*}{Accuracy } & \multirow{2}{*}{Train time }  & \multirow{2}{*}{Test time} & \multirow{2}{*}{Accuracy } & \multirow{2}{*}{Train time}  & \multirow{2}{*}{Test time} & \multicolumn{3}{c|}{TSVM} & \multicolumn{3}{c|}{LSTSVM} \\ \cline{8-10} \cline{11-13}
				& & & & & & & Accuracy & Train time & Test time & Accuracy & Train time & Test time \\
				\hline
				NDC-5K (5,000) & 87.4 & 0.029 & 0.003 & 86.77 &  0.683 & 0.005 & 86.77 & 0.1717 & 0.00015 & 86.57 & 0.0031 & 0.00012 \\
				NDC-10K (10,000) & 86.3 & 0.052 & 0.004 & 87.09 & 2.659 & 0.01 & 87.09 & 0.4897 & 0.00017 & 86.99 & 0.0055 & 0.00016 \\
				NDC-25K (25,000) & 85.76 & 0.119 & 0.008 & 86.47 & 16.444 & 0.0107 & 86.47 & 2.559 & 0.00036  & 86.35 & 0.0137 & 0.0005 \\
				NDC-50K (50,000) & 86.24 & 0.235 & 0.017 & 86.28 & 67.4271 & 0.0214 & 86.28 & 10.997 & 0.00059 & 86.20 & 0.0298 & 0.00091 \\
				NDC-1l (100,000) & 86.02 & 0.466 & 0.033 & \tnote{a} & \tnote{a}  & \tnote{a} & 85.75 & 54.609 & 0.0010 & 85.94 & 0.0512 & 0.0016 \\
				NDC-5l (500,000) & 85.93 & 2.275 & 0.15 & \tnote{a} & \tnote{a} & \tnote{a} & \tnote{a} & \tnote{a} & \tnote{a} & 85.92 & 0.2356 & 0.006 \\ 
				NDC-1m (1,000,000) & 86.20 & 4.547 & 0.295 & \tnote{a} & \tnote{a} & \tnote{a} & \tnote{a} & \tnote{a} & \tnote{a} & 86.25 & 0.4457 & 0.0117 \\
				NDC-2m (2,000,000) & 85.98 & 9.274 & 0.597 & \tnote{a} & \tnote{a} & \tnote{a} & \tnote{a} & \tnote{a} & \tnote{a} & 86.02 & 0.8630 & 0.0232 \\
				NDC-5m (5,000,000) & 86.00 & 22.899 & 1.494 & \tnote{a} & \tnote{a} & \tnote{a} & \tnote{a} & \tnote{a} & \tnote{a} & 86.05 & 2.1975 & 0.128 \\
				\hline
		\end{tabular}}
		\begin{tablenotes}
			\item[a] The experiment ran out of memory.
		\end{tablenotes}
	\end{threeparttable}
	\caption{A performance comparison between LIBLINEAR, LightTwinSVM, and LIBTwinSVM in terms of learning and prediction speed. Train and test time measured in seconds.}
	\label{tab:1}
\end{sidewaystable*}
	
%\end{table}

% Acknowledgements should go at the end, before appendices and references

\acks{This research work was carried out at the Machine Learning and Text Mining Lab of IranDoc Institution. We would like to thank the Director of the IranDoc Institution for providing us research facilities.}


%\appendix
%\section*{Appendix A.}
%\label{app:theorem}
%
%% Note: in this sample, the section number is hard-coded in. Following
%% proper LaTeX conventions, it should properly be coded as a reference:
%
%%In this appendix we prove the following theorem from
%%Section~\ref{sec:textree-generalization}:
%
%In this appendix we prove the following theorem from
%Section~6.2:
%
%\noindent
%{\bf Theorem} {\it Let $u,v,w$ be discrete variables such that $v, w$ do
%not co-occur with $u$ (i.e., $u\neq0\;\Rightarrow \;v=w=0$ in a given
%dataset $\dataset$). Let $N_{v0},N_{w0}$ be the number of data points for
%which $v=0, w=0$ respectively, and let $I_{uv},I_{uw}$ be the
%respective empirical mutual information values based on the sample
%$\dataset$. Then
%\[
%	N_{v0} \;>\; N_{w0}\;\;\Rightarrow\;\;I_{uv} \;\leq\;I_{uw}
%\]
%with equality only if $u$ is identically 0.} \hfill\BlackBox
%
%\noindent
%{\bf Proof}. We use the notation:
%\[
%P_v(i) \;=\;\frac{N_v^i}{N},\;\;\;i \neq 0;\;\;\;
%P_{v0}\;\equiv\;P_v(0)\; = \;1 - \sum_{i\neq 0}P_v(i).
%\]
%These values represent the (empirical) probabilities of $v$
%taking value $i\neq 0$ and 0 respectively.  Entropies will be denoted
%by $H$. We aim to show that $\fracpartial{I_{uv}}{P_{v0}} < 0$....\\
%
%{\noindent \em Remainder omitted in this sample. See http://www.jmlr.org/papers/ for full paper.}

\appendix
\section*{Appendix A. Additional Experiments}
\label{app:aexp}
In this section, we further analyze the efficiency of the LIBTwinSVM library in comparison with other popular implementations. First, we compare the classification performance of the library with other implementations on benchmark datasets. Second, we analyze the memory consumption of the LIBTwinSVM library to validate the effectiveness of our implementation. These two experiments were carried out on a PC with Intel Core i7 6700K CPU, 32.0 GB of RAM, and Ubuntu 16.04.6 LTS. The version of our software and the baseline libraries are shown in Table \ref{tab:6}.

\begin{table}[H]
	\small
	\centering
	\begin{tabular}{l c c c c c}
		\hline
		& LIBLINEAR & LIBSVM & ThunderSVM &  LightTwinSVM & LIBTwinSVM \\
		\hline
		Version & v2.30 & v3.23 & v0.3.3 & v0.6.0 & v0.3.0 \\
		\hline
	\end{tabular}
	\caption{The version of our software and the baseline libraries.}
	\label{tab:6}
\end{table}

\subsection*{A.1 Benchmark datasets}
To show the classification performance of the LIBTwinSVM library, we conducted experiments on benchmark datasets from the UCI\footnote{http://archive.ics.uci.edu/ml/datasets} machine learning repository. The feature values of all the datasets were normalized in the range $[0,1]$. For multi-classification, we chose One-vs-One scheme as it is often used in SVM packages. The characteristics of UCI benchmark datasets are shown in the Table \ref{tab:2}.

\begin{table}[!t]
	\small
	\centering
	\begin{tabular}{l c c c}
		\hline
		Datasets & \#Samples & \#Features & \#Class \\
		\hline
		%\textit{Small datasets} & & & \\
		Iris         &    150    &   4    &  3 \\
		Hepatitis    &    155    &   19   &  2 \\
		Monk3        &    554    &   6    &  2 \\
		Australian   &    690    &   14   &  2 \\ 
		% Bupa-Liver   &    345    &   6    &  2 \\
		%Ionsphere    &    351    &   34   &  2 \\
		
		%WPBC         &    198    &   30   &  2 \\
	
		%Parkinson    &    195    &   22   &  2 \\
		%Breast Cancer &    569    &   32   &  2 \\
		%Planning Relax &    182    &   12   &  2 \\
	    %Wine         &    178    &   13   &  3 \\  
	    %Balance      &    625    &   4    &  3 \\
		Vehicle      &    846    &   18   &  4 \\
		
        %Vowel        &    625    &   13   &  11 \\
        Optdigits    &    1797    &   64   &  10 \\
        Landsat      &    2000    &   36   &  6 \\
        % & & & \\
        %\textit{Medium datasets} & & & \\
        Satimage    &    6435   &   36   &  6 \\
        Mushrooms    &    8124    &   112   &  2 \\
        Pendigits    &    10992    &   16   &  10 \\
        % & & & \\
        %\textit{Large datasets} & & & \\
         w3a & 49749 & 300 &  2 \\
        
		\hline
	\end{tabular}
	\caption{The characteristics of UCI benchmark datasets}
	\label{tab:2}
\end{table}

The classification performance of SVM and TSVM classifiers depends heavily on the choice of hyper-parameters. The optimal value of $c_{1}$ and $c_{2}$ parameters was selected from the set $\{2^{i} \mid i=-5,-4,\dots,5 \}$. In this experiment, the grid search method is used to find the optimal values of hyper-parameters. We evaluated the classification performance of the classifiers using 5-fold cross-validation. Moreover, a linear kernel is used and the stopping criteria of the clipDCD optimizer is set to $10^{-5}$.

Table \ref{tab:3} shows the accuracy comparison of LIBTwinSVM and other implementations with a linear kernel. From the Table \ref{tab:3}, it can be seen that the LIBTwinSVM library outperforms popular SVM packages (i.e. LIBLINEAR , LIBSVM and ThunderSVM). For instance, the accuracy of the LIBTwinSVM (TSVM) is about 14\% higher than that of LIBLINEAR on Satimage dataset. On the mentioned dataset, the training time of the LIBTwinSVM (TSVM) is 5.8 times faster than that of LIBLINEAR. The accuracy advantage of TSVM is due to the fundamental difference in the central idea of SVM and TSVM. As stated previously, TSVM classifier does classification by using two non-parallel hyperplanes. Also, note that the accuracy of LightTwinSVM and LIBTwinSVM (TSVM) is the same. Because they use the same classifier for solving a classification task. However, the classification accuracy of LIBTwinSVM (LSTSVM) is slightly better than LightTwinSVM and LIBTwinSVM (TSVM) on some datasets. Even though the central idea of LSTSVM estimator is the same as TSVM estimator, the solution of LSTSVM estimator may be different for a classification problem.

In addition to the experiment with a linear kernel, we conducted an experiment on benchmark datasets with RBF kernel. Table \ref{tab:5} shows the accuracy comparison of LIBTwinSVM and other implementations with RBF kernel. The parameter of RBF kernel was chosen from the set $\{2^{i} \mid i=-15,-4,\dots,2 \}$. The classification results of LIBTwinSVM with RBF kernel is comparable or slightly better then the implementations of SVM. Moreover, the training and prediction speed of the LIBTwinSVM (TSVM) is substantially faster than that of LightTwinSVM. For example,  the training speed of the LIBTwinSVM (TSVM) is about 12 times faster than LightTwinSVM for Pendigits dataset. This validates the improvements in the implementation of LIBTwinSVM library.

In summary, the results on benchmark datasets indicate that the LIBTwinSVM library is suitable for solving real-world classification problems.

\begin{sidewaystable*}
	%\begin{table}[!t]
	\small
	\centering
	\begin{threeparttable}
	\begin{tabular}{|l | c | c | c | c | c | c|}
		\hline
		\multirow{3}{*}{Datasets} & LIBLINEAR & LIBSVM & ThunderSVM & LightTwinSVM & LIBTwinSVM (TSVM) & LIBTwinSVM (LSTSVM) \\ \cline{2-3} \cline{4-5} \cline{6-7}
		& Tr/Te (s) & Tr/Te (s) & Tr/Te (s) & Tr/Te (s) & Tr/Te (s) & Tr/Te (s) \\ \cline{2-3} \cline{4-5} \cline{6-7}
		& $C$ & $C$ & $C$ & ($C_{1},C_{2}$) & ($C_{1},C_{2}$) & ($C_{1},C_{2}$) \\  
		\hline
		Iris & 95.33$\pm$0.03 & 96.67$\pm$0.03 & 96.67$\pm$0.03 & 98.00$\pm$2.67 & 98.00$\pm$2.67 & \textbf{98.67$\pm$1.63} \\
		& 0.005/0.00004 & 0.001/0.0003 & 0.005/0.0001 & 0.006/0.002 & 0.005/0.002 & 0.005/0.002 \\
		& $2^{2}$ & $2^{-1}$ & $2^{-1}$ & $(2^{-5}, 2^{-3})$ & $(2^{-5}, 2^{-3})$ & $(2^{-1}, 2^{-2})$ \\
		\hline
		Hepatitis & 82.58$\pm$0.02 & 78.71$\pm$0.03 & 78.71$\pm$0.03 & 80.65$\pm$6.45 & 80.65$\pm$6.45 & \textbf{83.23$\pm$5.55} \\
		& 0.018/0.0002 & 0.003/0.0004 & 0.007/0.0002 & 0.006/0.0003 & 0.0017/0.0002 & 0.0011/0.0001 \\
		& $2^{5}$ & $2^{-5}$ & $2^{-5}$ & $(2^{-5}, 2^{-5})$ & $(2^{-5}, 2^{-5})$ & $(2^{-5}, 2^{-3})$ \\
		\hline
		Monk3 & 78.88$\pm$0.04 & 80.14$\pm$0.03 & 80.14$\pm$0.03 & \textbf{86.10$\pm$2.75} & \textbf{86.10$\pm$2.75} & 85.74$\pm$2.26 \\
		& 0.013/0.0002 & 0.02/0.003 & 0.026/0.0004 & 0.016/0.0009 & 0.002/0.0002 & 0.001/0.0001 \\
		& $2^{5}$ & $2^{-5}$ & $2^{-5}$ & $(2^{2}, 2^{0})$ & $(2^{2}, 2^{0})$ & $(2^{1}, 2^{-1})$ \\
		\hline
		Australian & 86.81 $\pm$ 0.04  & 85.51 $\pm$  0.03  &  85.51 $\pm$  0.03 &  87.10 $\pm$ 2.40 &   87.10 $\pm$ 2.40 & \textbf{87.68} $\pm$ \textbf{2.50} \\
		& 0.023/0.00007  & 0.01/0.002 & 0.026/0.0008 &  0.029/0.001 & 0.004/0.0002 & 0.0004/0.00006 \\
		& $2^{4}$  & $2^{-5}$ & $2^{4}$ &  ($2^{-3},2^{-5},2^{0}$) & ($2^{-3},2^{-5},2^{0}$) & ($2^{-3},2^{-4},2^{0}$) \\
		\hline
		%Bupa-Liver & 68.41 $\pm$ 0.07  & \textbf{70.43} $\pm$ \textbf{0.03} &  70.43 $\pm$  0.05 &  69.57 $\pm$ 2.43 & 69.86 $\pm$ 6.57 & \textbf{70.43} $\pm$ \textbf{2.98} \\
		%Ionsphere  & 88.03 $\pm$ 0.03  &  88.32 $\pm$ 0.04 &  89.74 $\pm$  0.02 &  89.75 $\pm$ 2.25 &  \textbf{90.02} $\pm$ \textbf{3.02} & 89.7 $\pm$ 5.58 \\
		%WPBC   &  81.82 $\pm$ 0.05  &  81.31 $\pm$ 0.03 &  81.82 $\pm$  0.09 &  77.31 $\pm$ 4.32 &  77.81 $\pm$ 4.75 & \textbf{83.31} $\pm$ \textbf{3.57} \\
		%Parkinson & 76.92 $\pm$ 0.04  & 87.18 $\pm$ 0.03 &  88.72 $\pm$  0.02 &  87.18 $\pm$ 3.63 & 86.67 $\pm$ 5.94 &  \textbf{89.23} $\pm$ \textbf{4.97} \\
		%Breast Cancer  & 97.14 $\pm$ 0.02  & 96.85 $\pm$ 0.01 &  97.00 $\pm$  0.01 & 96.57 $\pm$ 1.94 & \textbf{97.28} $\pm$ \textbf{1.14}  &  97.14 $\pm$ 1.10 \\
		%Planning Relax  & 70.80 $\pm$ 0.02  & 71.43 $\pm$ 0.01 &  71.43 $\pm$  0.01 & 70.90 $\pm$ 4.90 & 72.00 $\pm$ 3.00  &  \textbf{72.04} $\pm$ \textbf{6.55} \\
		%Wine & 98.31 $\pm$ 0.01  &  98.88 $\pm$ 0.01  &  98.88 $\pm$  0.01 & 96.65 $\pm$ 3.24 &  \textbf{99.43} $\pm$ \textbf{1.14} &  \textbf{99.43} $\pm$ \textbf{1.14} \\
		%Balance & 89.12 $\pm$ 0.02 & \textbf{91.68} $\pm$ \textbf{0.01}  &  91.68 $\pm$  0.02 &  90.24 $\pm$ 1.47 &  91.52 $\pm$ 0.82 &  87.68 $\pm$ 2.71 \\
		
		Vehicle & 70.92$\pm$0.05 & 79.79$\pm$0.04 & 79.79$\pm$0.04 & \textbf{81.33$\pm$2.86} & \textbf{81.33$\pm$2.86} & 80.74$\pm$2.70 \\
		& 0.12/0.00007 & 0.76/0.003 & 3.35/0.001 & 0.03/0.02 & 0.008/0.03 & 0.006/0.03 \\ 
		& $2^{-2}$ & $2^{-4}$ & $2^{-4}$ & $(2^{-2}, 2^{0})$ & $(2^{-2}, 2^{0})$ & $(2^{-3}, 2^{-2})$ \\ 
		\hline
		Optdigits & 95.88$\pm$0.01 & 98.00$\pm$0.01 & 98.05$\pm$0.01 & 97.16$\pm$0.48 & 97.16$\pm$0.48 & \textbf{98.16$\pm$0.67} \\
		& 0.10/0.0001 & 0.06/0.02 & 0.67/0.01 & 0.26/0.20 & 0.07/0.24 & 0.13/0.25 \\ 
		& $2^{-5}$ & $2^{-5}$ & $2^{-5}$ & $(2^{-4}, 2^{-5})$ & $(2^{-4}, 2^{-5})$ & $(2^{-2}, 2^{-3})$ \\
		\hline
	    Landsat & 66.55$\pm$0.06 & 84.55$\pm$0.02 & 84.55$\pm$0.02 & 84.75$\pm$0.79 & 84.75$\pm$0.79 & \textbf{84.90$\pm$0.87} \\ 
	    & 0.45/0.0001 & 0.39/0.01 & 97.20/0.015 & 0.36/0.10 & 0.05/0.13 & 0.04/0.13 \\ 
        & $2^{2}$ & $2^{-5}$ & $2^{-5}$ & $(2^{-2}, 2^{0})$ & $(2^{-2}, 2^{0})$ & $(2^{-1}, 2^{0})$ \\ 
		\hline
		%Vowel   & 57.88 $\pm$ 0.03 & 83.33 $\pm$ 0.01  &  \textbf{83.43} $\pm$  \textbf{0.03} &  77.37 $\pm$ 3.25 & 75.96 $\pm$ 3.76 &  76.67 $\pm$ 1.17  \\
		Satimage & 71.93$\pm$0.07 & \multirow{3}{*}{\tnote{b}} & \multirow{3}{*}{\tnote{b}} & 86.45$\pm$0.88 & \textbf{86.45$\pm$0.88} & 85.66$\pm$1.14 \\
		& 1.63/0.002 &  &  & 1.72/0.22 & 0.28/0.29 & 0.021/0.29 \\ 
		& $2^{-4}$ &  &  & $(2^{-1}, 2^{0})$ & $(2^{-1}, 2^{0})$ & $(2^{0}, 2^{1})$ \\ 
		\hline
		Mushrooms & \textbf{100.00$\pm$0.00} & \multirow{3}{*}{\tnote{b}} & \multirow{3}{*}{\tnote{b}} & \textbf{100.00$\pm$0.00} & \textbf{100.00$\pm$0.00} & \textbf{100.00$\pm$0.00} \\
		& 0.021/0.0002 &  &  & 1.04/0.014 & 0.17/0.0002 & 0.012/0.0001 \\
		& $2^{-4}$ &  &  & $(2^{-5}, 2^{-5})$ & $(2^{-5}, 2^{-5})$ & $(2^{-5}, 2^{-5})$ \\
		\hline
		Pendigits & 88.44$\pm$0.01 & \multirow{3}{*}{\tnote{b}} & \multirow{3}{*}{\tnote{b}} & 96.86$\pm$0.53 & \textbf{96.86$\pm$0.53} & 96.73$\pm$0.41 \\ 
		& 1.52/0.0004 &  &  & 5.14/0.97 & 1.03/1.27 & 0.037/1.28 \\
		& $2^{0}$ &  &  & $(2^{-2}, 2^{-1})$ & $(2^{-2}, 2^{-1})$ & $(2^{0}, 2^{0})$ \\  
		\hline
		w3a & \textbf{98.67$\pm$0.01} & \multirow{3}{*}{\tnote{b}} & \multirow{3}{*}{\tnote{b}} & \multirow{3}{*}{\tnote{a}} & 98.31$\pm$0.12 & 98.55$\pm$0.06 \\
		& 0.78/0.004 &  & &  & 13.24/0.002 & 0.15/0.001 \\
		& $2^{2}$ &  &  &  & $(2^{-2}, 2^{-1})$ & $(2^{-3}, 2^{-2})$ \\  
		\hline
		Mean accuracy & 85.08 & - & - & 90.61  & 90.61  & \textbf{90.91} \\
		\hline
	\end{tabular}
    \begin{tablenotes}
    	\item[a] The experiment ran out of memory.
    	\item[b] We have not performed the experiment as the computing time was very high.
    \end{tablenotes}
    \end{threeparttable}
	\caption{The accuracy comparison of LIBTwinSVM and other implementations with linear kernel. Tr/Te stands for train and test time.}
	\label{tab:3}
	%\end{table}
\end{sidewaystable*}

\begin{sidewaystable*}
	%\begin{table}[!t]
	\small
	\centering
	\begin{threeparttable}
		\begin{tabular}{|l | c | c | c | c | c |}
			\hline
			\multirow{3}{*}{Datasets (b)} & LIBSVM & ThunderSVM & LightTwinSVM & LIBTwinSVM (TSVM) & LIBTwinSVM (LSTSVM) \\ \cline{2-3} \cline{3-4} \cline{5-6}
			& Tr/Te (s) & Tr/Te (s) & Tr/Te (s) & Tr/Te (s) & Tr/Te (s) \\ \cline{2-3} \cline{3-4} \cline{5-6}
			& $(C, \sigma)$ & $(C, \sigma)$ & ($C_{1},C_{2}, \sigma$) & ($C_{1},C_{2}, \sigma$) & ($C_{1},C_{2}, \sigma$) \\  
			\hline
			Iris & 96.67$\pm$0.04 & 96.67$\pm$0.04 & 98.00$\pm$1.63 & 98.00$\pm$1.63 & \textbf{98.67$\pm$1.63} \\ 
			& 0.0008/0.0001 & 0.01/0.0005 & 0.13/0.004 & 0.029/0.004 & 0.13/0.004 \\ 
			& $(2^{1}, 2^{-4})$ & $(2^{1}, 2^{-4})$ & $(2^{-3}, 2^{-2}, 2^{-4})$ & $(2^{-3}, 2^{-2}, 2^{-4})$ & $(2^{2}, 2^{2}, 2^{-4})$ \\
			\hline
			Hepatitis & \textbf{86.45$\pm$0.03} & \textbf{86.45$\pm$0.03} & 85.81$\pm$2.58 & 85.81$\pm$2.58 & 85.81$\pm$2.58 \\ 
			& 0.004/0.0008 & 0.001/0.0002 & 0.008/0.001 & 0.009/0.0003 & 0.007/0.0003 \\ 
			& $(2^{3}, 2^{-7})$ & $(2^{3}, 2^{-7})$ & $(2^{-4}, 2^{0}, 2^{-13})$ & $(2^{-4}, 2^{0}, 2^{-13})$ & $(2^{-5}, 2^{0}, 2^{-7})$ \\ 
			\hline
			Monk3 & 97.11$\pm$0.02 & 97.11$\pm$0.02 & 97.65$\pm$0.72 & 97.65$\pm$0.72 & \textbf{98.38$\pm$0.36} \\
			& 0.015/0.001 & 0.012/0.0004 & 0.09/0.007 & 0.10/0.002 & 0.12/0.006 \\
			& $(2^{5}, 2^{-4})$ & $(2^{5}, 2^{-4})$ & $(2^{-5}, 2^{-5}, 2^{-3})$ & $(2^{-5}, 2^{-5}, 2^{-3})$ & $(2^{-5}, 2^{-2}, 2^{-3})$ \\
			\hline 
			Australian & 87.10$\pm$0.03 & 87.10$\pm$0.03 & \textbf{87.54$\pm$2.98} & \textbf{87.54$\pm$2.98} & 87.39$\pm$2.99 \\
			& 0.02/0.004 & 0.36/0.04 & 0.27/0.01 & 0.17/0.005 & 0.13/0.005 \\ 
			& $(2^{-4}, 2^{-4})$ & $(2^{-4}, 2^{-4})$ & $(2^{2}, 2^{-1}, 2^{-9})$ & $(2^{2}, 2^{-1}, 2^{-9})$ & $(2^{-5}, 2^{-5}, 2^{-7})$ \\ 
			\hline
			Vehicle & 82.51$\pm$0.04 & 82.51$\pm$0.04 & 83.46$\pm$2.86 & 83.46$\pm$2.86 & \textbf{84.99}$\pm$2.29 \\
			& 0.04/0.008 & 0.04/0.002 & 0.24/0.07 & 0.11/0.07 & 0.11/0.07 \\
			& $(2^{5}, 2^{-7})$ & $(2^{5}, 2^{-7})$ & $(2^{-3}, 2^{-3}, 2^{-5})$ & $(2^{-3}, 2^{-3}, 2^{-5})$ & $(2^{-5}, 2^{-4}, 2^{-6})$ \\
			\hline
			Optdigits & 99.28$\pm$0.00 & 99.28$\pm$0.00 & 98.89$\pm$1.00 & 98.89$\pm$1.00 & \textbf{99.50$\pm$0.59} \\
			& 0.36/0.045 & 0.83/0.015 & 1.64/1.31 & 0.88/1.48 & 0.65/1.47 \\
			& $(2^{1}, 2^{-10})$ & $(2^{1}, 2^{-10})$ & $(2^{-4}, 2^{-4}, 2^{-13})$ & $(2^{-4}, 2^{-4}, 2^{-13})$ & $(2^{0}, 2^{-2}, 2^{-12})$ \\ 
			\hline 
			Landsat & 88.10$\pm$0.02 & 88.10$\pm$0.02 & 88.05$\pm$1.48 & 88.05$\pm$1.48 & \textbf{88.75$\pm$1.52} \\
			& 0.34/0.04 & 0.30/0.006 & 0.97/0.58 & 0.62/0.63 & 0.62/0.64 \\
			& $(2^{4}, 2^{-7})$ & $(2^{4}, 2^{-7})$ & $(2^{-2}, 2^{-1}, 2^{-7})$ & $(2^{-2}, 2^{-1}, 2^{-7})$ & $(2^{3}, 2^{3}, 2^{-5})$ \\
			\hline
			Satimage (0.25) & 90.69$\pm$0.01 & 90.69$\pm$0.01 & 89.95$\pm$0.51 & 89.95$\pm$0.51 & \textbf{90.78$\pm$0.69} \\ 
			& 3.22/0.42 & 1.91/0.09 & 12.76/3.48 & 1.006/1.91 & 0.69/1.92 \\
			& $(2^{5}, 2^{-7})$ & $(2^{5}, 2^{-7})$ & $(2^{-3}, 2^{-3}, 2^{-5})$ & $(2^{-3}, 2^{-3}, 2^{-5})$ & $(2^{-4}, 2^{-4}, 2^{-5})$ \\ 
			\hline
			Mushrooms (0.25) & \textbf{100.00}$\pm$0.00 & \textbf{100.00}$\pm$0.00 & 99.79$\pm$0.08 & 99.79$\pm$0.08 & 99.99$\pm$0.02 \\ 
			& 0.69/0.15 & 0.59/0.043 & 2.43/0.26 & 1.71/0.16 & 0.74/0.17 \\ 
			& $(2^{0}, 2^{-4})$ & $(2^{0}, 2^{-4})$ & $(2^{-5}, 2^{-5}, 2^{-2})$ & $(2^{-5}, 2^{-5}, 2^{-2})$ & $(2^{-5}, 2^{-1}, 2^{-2})$ \\ 
			\hline
			Pendigits (0.25) & 99.47$\pm$0.01 & 99.47$\pm$0.01 & 99.18$\pm$0.24 & 99.18$\pm$0.24 & \textbf{99.49$\pm$0.13} \\ 
			& 7.70/1.009 & 4.02/0.22 & 35.05/17.614284 & 2.89/4.26 & 2.05/4.20 \\
			& $(2^{5}, 2^{-7})$ & $(2^{5}, 2^{-7})$ & $(2^{-3}, 2^{-2}, 2^{-6})$ & $(2^{-3}, 2^{-2}, 2^{-6})$ & $(2^{-3}, 2^{-3}, 2^{-5})$ \\
			\hline
			w3a (0.05) & \textbf{98.90$\pm$0.01} & \textbf{98.90$\pm$0.01} & \multirow{3}{*}{\tnote{a}} & 98.56$\pm$0.10 & 98.57$\pm$0.06 \\ 
		    & 9.13/1.87 & 20.39/1.28 &  & 47.07/0.90 & 3.36/0.91 \\
			& $(2^{3}, 2^{-10})$ & $(2^{3}, 2^{-10})$ &  & $(2^{-1}, 2^{-1}, 2^{-14})$ & $(2^{-5}, 2^{-3}, 2^{-14})$ \\ 
			\hline
			Mean accuracy & 93.29 & 93.29 & - & 93.32 & \textbf{93.84} \\
			\hline
		\end{tabular}
		\begin{tablenotes}
			\item[a] The experiment ran out of memory.
			\item[b] The value of the rectangular kernel for LightTwinSVM and LIBTwinSVM.
			%\item[b] We have not perform the experiment as the computing time was very high.
		\end{tablenotes}
	\end{threeparttable}
	\caption{The accuracy comparison of LIBTwinSVM and other implementations with RBF kernel}
	\label{tab:5}
	%\end{table}
\end{sidewaystable*}

\subsection*{A.2 Analysis of memory consumption}
Here, we analyze the memory consumption of the LIBTwinSVM library in comparison with other implementations. For this experiment, the NDC datasets were used. To experiment with LIBLINEAR, LIBSVM, and ThunderSVM, their Python bindings were employed. Our procedure for measuring the peak memory consumption is (1) we load a dataset. (2) we train an estimator using \texttt{fit()} method on the entire dataset. We note that the second step of the procedure is considered for the analysis of memory consumption. Because the \texttt{fit()} method is the most resource-intensive part in the code. We used the Linux \texttt{time} command to report the peak memory usage. 

Table \ref{tab:4} shows the peak memory consumption of the LIBTwinSVM library and other implementations with linear kernel\footnote{RBF kernel is computationally expensive and its memory usage is quite high. Therefore, we analyzed the memory consumption with linear kernel.}. Thanks to the improvements in the implementation of the LIBTwinSVM, the memory consumption of the LIBTwinSVM estimators is much lower than that of the LightTwinSVM program. However, the memory usage of the LIBTwinSVM's TSVM estimator is higher than the implementations of SVM. This is because TSVM needs to store two matrices of size $m_1 \times m_1$ and $m_2 \times m_2$ in order to solve two dual optimization problems (where $m_1+m_2=m$ and $m$ denotes the number of samples.). On the other hand, the memory usage of the LIBTwinSVM's LSTSVM estimator is significantly lower than LIBLINEAR for 50K samples and more.

	\begin{table}[!t]
	\small
	\centering
	\begin{threeparttable}
		\resizebox{\textwidth}{!}{\begin{tabular}{|c |c |c |c |c |c |c |}
				\hline
				Datasets & LIBLINEAR & LIBSVM & ThunderSVM & LightTwinSVM & LIBTwinSVM (TSVM) & LIBTwinSVM (LSTSVM) \\
				\hline
				NDC-5K (5,000) & 47.09 & 135.85 &  137.57 & 343.20 & 170.41 & 96.51 \\
				NDC-10K (10,000) & 64.33 & 164.625 & 189.07 & 982.62 & 392.96 & 102.03 \\
				NDC-25K (25,000) & 117.83 & 218.71 & 340.5 & 6254.75 & 1934.89 & 117.66 \\
				NDC-50K (50,000) & 206.86 & 308.75 & 593.37 & 24659.29 & 7387.24 & 141.80 \\
				NDC-1l (100,000) & 384.34 & \tnote{b} & 1097.46 & \tnote{a} & 29242.50 & 191.87 \\
				NDC-5l (500,000) & 1799.64 & \tnote{b} & \tnote{b} & \tnote{a} & \tnote{a} & 591.40 \\
				NDC-1m (1,000,000) & 3571.56  & \tnote{b} & \tnote{b} & \tnote{a} & \tnote{a} & 1143.37 \\
				NDC-2m (2,000,000) & 7122.47 & \tnote{b} & \tnote{b} & \tnote{a} & \tnote{a} & 2193.11 \\
				NDC-5m (5,000,000)& 17727.84 & \tnote{b} & \tnote{b} & \tnote{a} & \tnote{a} & 5110.21 \\
				\hline
		\end{tabular}}
		\begin{tablenotes}
			{\small \item[a] The experiment ran out of memory.
			\item[b] We have not performed the experiment as the computing time was very high.}
		\end{tablenotes}
	\end{threeparttable}
	\caption{The peak memory consumption of the LIBTwinSVM library and other implementations. The reported numbers are in megabytes.}
	\label{tab:4}
	\end{table}
%\end{sidewaystable*}

\vskip 0.2in
\bibliography{sample}
\end{document}